\pdfoutput=1
\documentclass[11pt]{article}
\usepackage[letterpaper,margin=1in]{geometry}
\usepackage[numbers,sort&compress]{natbib}
\usepackage{times}
\usepackage[utf8]{inputenc}
\usepackage[T1]{fontenc}
\usepackage{hyperref}
\usepackage{url}
\usepackage{booktabs}
\usepackage{amsfonts,amsmath,amssymb,amsthm}
\usepackage{microtype}
\usepackage{xcolor}
\usepackage{colortbl}
\usepackage{graphicx}
\usepackage{enumitem}
\usepackage{caption}

\definecolor{pos}{HTML}{00703C}
\definecolor{neg}{HTML}{B00020}
\definecolor{calblue}{HTML}{0072B2}
\definecolor{keyrow}{HTML}{E6F0FA}
\hypersetup{colorlinks=true,linkcolor=calblue,citecolor=calblue,urlcolor=calblue}
\setlist{leftmargin=1.4em,itemsep=1pt,topsep=2pt}

\newcommand{\pplT}{\mathrm{PPL}_T}
\newcommand{\para}[1]{\paragraph{#1}}
\newcommand{\mns}{\ensuremath{-}}
\newcommand{\pls}{\ensuremath{+}}

\title{There Is No Free Loop: When Does Looping a Trained Model Pay for Its Inference Compute?}

\author{Rhea Sanjay Sukthanker \qquad Pashmina Cameron\\
Microsoft Applied Sciences Group\\[6pt]
\texttt{rsukthanker@microsoft.com} \qquad
\texttt{pcameron@microsoft.com}}
\date{}

\begin{document}
\maketitle

\begin{abstract}
Looping re-applies a block of layers so that each token passes through the block more than once. Adding a loop to a trained model is attractive because it leaves the parameter count, and hence the disk size, the memory footprint and the training cost unchanged. Unfortunately, every generated token pays for the loop. Re-applying $m$ of $N$ layers makes each token about $(N{+}m)/N$ times as expensive. Loops that a model learns during large-scale pretraining justify this price, but whether loops added after pretraining justify this cost remains unclear. We compare a looped model with two simple baselines. We first compare it with the same model trained identically without the loop (equal parameters). Secondly, we compare it with the best unlooped use of the same inference floating-point operations (equal FLOPs). Across Qwen3 and Qwen3.5 models from 0.8B to 9B parameters, only the looped 0.8B model beats both baselines. At 4B and 9B, the looped model does not beat the first baseline, i.e.\ the same model trained without the loop. This is not because the benchmarks are too easy, since on the same benchmarks the unlooped 9B model still scores up to 13 points higher than the 4B model, showing that there is ample room for improvement. Introducing loops in an already pretrained model therefore only pays for small models with a fixed parameter budget.
\end{abstract}

% =====================================================================================================
\section{Introduction}

A transformer model normally passes a token through each of its layers once. A \emph{loop} passes the token through a block of layers several times with the same weights, so the model spends more computation per token without storing more parameters. When a model learns its loop with a large pretraining budget, this pays off. Recurrent-depth models pretrained on hundreds of billions of tokens turn extra iterations into accuracy~\cite{geiping2025scaling}, and looped models pretrained on 7.7T tokens compete with dense models several times larger~\cite{zhu2025ouro}. However, few practitioners can afford to pretrain at that scale. A cheaper route introduces loops into a model that already exists, either without any training~\cite{chen2026training} or during the short post-training that turns a base model into a reasoning model~\cite{tah2026}. Such a loop does not change the model's disk size, memory footprint and training cost.

Unfortunately, every generated token pays for the loop. Re-applying $m$ of $N$ layers makes each token cost about $(N{+}m)/N$ as much, for as long as the model serves requests. The same compute has other uses. It buys the unlooped model longer answers, several sampled answers, or a move to a larger model if memory footprint is not a concern. A loop that beats its starting model is therefore not useful, unless it also beats these alternatives.

\begin{figure}[t]
\centering
\includegraphics[width=\linewidth]{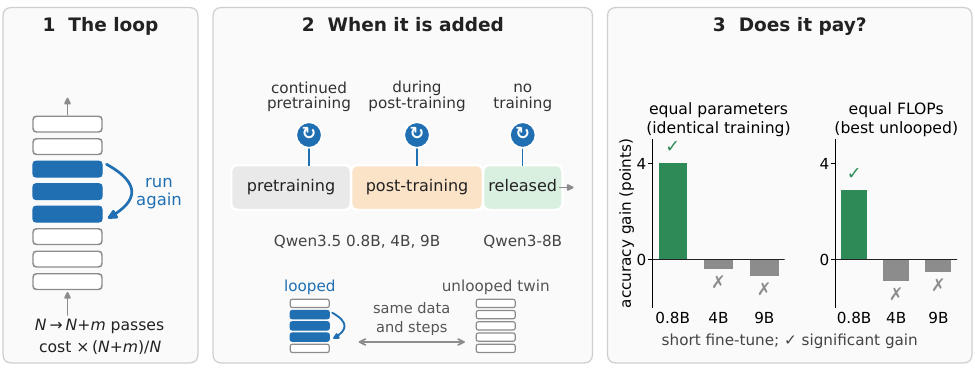}
\caption{\textbf{Testing a loop twice.} (1)~A loop re-runs $m$ of a model's $N$ layers for every token. (2)~We add loops at three points in a model's life cycle, and each looped model has a twin trained on the same data for the same steps. (3)~Bars show the looped model's accuracy gain after the short fine-tune, over its twin (left) and over the best unlooped evaluation of equal FLOPs (right). Only the 0.8B loop beats both.}
\label{fig:overview}
\end{figure}

Our key insight is that a loop introduced after training should pass two tests, in a fixed order. The first compares models at \emph{equal parameters} and asks whether the loop beats the same model trained identically without it. The second compares them at \emph{equal FLOPs}, i.e.\ equal inference floating-point operations, and asks whether the loop beats the best unlooped use of the same compute. A loop that fails the first test never reaches the second (Figure~\ref{fig:overview}). We test an assumption implicit in prior work on looping trained models, namely that a loop adds useful depth and should therefore pass both tests. We study training-free loops in a released reasoning model (Qwen3-8B), and loops that Qwen3.5 base models of 0.8B, 4B and 9B parameters learn while post-training turns them into reasoning models.

The hypothesis holds only for the smallest model. The 0.8B loop passes both tests, and this continues to hold when we strengthen both models in the same two ways. We fine-tune each model on three times as much data, and we decode each with \emph{budget forcing}~\cite{muennighoff2025s1}, a rule that ends the reasoning once a token limit is reached, which we place shortly before the full budget so that the model always writes an answer. At 4B and 9B the loop fails the first test, even though the 9B model scores up to 13 points higher than the 4B model on the same benchmarks. Hence, this is not an effect of the benchmarks saturating or being too easy. The key contributions of our work are:
\begin{itemize}[leftmargin=*,itemsep=1pt,topsep=2pt]
\item We propose a two-test protocol for loops added to trained models, which gives budget forcing to both models as a baseline that costs no extra compute (\S\ref{sec:protocol}).
\item We show that such loops pay only at small scale, and that larger models gain nothing from them even when we ignore the extra compute introduced (\S\ref{sec:post}).
\item We measure the serving cost and derive a practical rule. A loop is worth its price only when the parameter budget cannot grow (\S\ref{sec:serving}).
\end{itemize}

% =====================================================================================================
\section{What do we know about loops so far?}
\label{sec:background}

\para{Loops learned from the start.} Reusing depth for recursive models to improve model quality while keeping the number of parameters constant is a widely explored idea. Residual networks behave like unrolled iterative estimators that refine one representation layer by layer~\cite{greff2017highway,jastrzebski2018residual}, and a deep weight-shared residual network is equivalent to a shallow recurrent network~\cite{liao2016bridging}. Neural GPUs learn algorithms by applying one block many times~\cite{kaiser2016neural}, and deep equilibrium models extend weight tying to infinite depth~\cite{bai2019deep}. Universal transformers apply one shared block repeatedly~\cite{dehghani2019universal}, and adaptive computation time and PonderNet learn how many steps each input needs~\cite{graves2016act,banino2021pondernet}. Looped transformers can emulate programs~\cite{giannou2023looped}, learn learning algorithms in context~\cite{yang2024looped} and generalise to longer inputs~\cite{fan2025looped}. When trained from scratch, a $k$-layer transformer looped $L$ times nearly matches a $kL$-layer model on synthetic reasoning tasks~\cite{saunshi2025reasoning}, and depth that grows only logarithmically with input length provably adds expressive power~\cite{merrill2025littledepth}. At language-model scale, recurrent-depth and looped models pretrained on 0.8--7.7T tokens benefit from extra recurrent iterations at test time, within the range of depths they are trained on~\cite{geiping2025scaling,zhu2025ouro}. Mixture-of-recursions models learn per-token recursion depth during pretraining~\cite{bae2025mor}, and mixture-of-experts universal transformers make shared-layer models competitive in language-model pretraining~\cite{csordas2024moeut}. Models can also spend extra sequential computation in a continuous latent space~\cite{hao2025coconut} or on filler tokens~\cite{pfau2024dot}. Hence recurrence pays when the model learns it with a large compute and data budget.

\begin{table}[t]
\centering
\caption{\textbf{Tokens spent learning the loop.} Summary of model sizes and number of tokens used by methods which train looped transformers. Shaded rows mark our settings, and bold numbers are our token budgets.}
\label{tab:budgets}
\small
\begin{tabular}{lll}
\toprule
Work & Model & Tokens \\
\midrule
Huginn~\cite{geiping2025scaling} & 3.5B, from scratch & 800B \\
Ouro~\cite{zhu2025ouro} & 1.4B / 2.6B, from scratch & 7.7T \\
Retrofitted recurrence~\cite{mcleish2025teaching} & 1B, converted & $\approx$50B \\
Closing the Loop~\cite{closingloop2026} & 1.7B, converted & 60--84B \\
TaH2~\cite{tah2026} & 1.7--8B, post-training & 3.4--16B \\
\midrule
\rowcolor{keyrow} Ours & 0.8--9B, post-training & $\le$\textbf{0.5B} \\
\rowcolor{keyrow} Ours & 0.8B / 4B, continued pretraining & $\le$\textbf{5.2B} \\
\rowcolor{keyrow} Ours & Qwen3-8B, conversion & \textbf{33--66M} \\
\bottomrule
\end{tabular}
\end{table}

\para{Converting a pretrained model.} Converting an existing model into a looped one also works, but every successful conversion so far spends tens of billions of tokens on learning the loop (Table~\ref{tab:budgets}). Relaxed recursive transformers turn pretrained models into layer-shared ones with per-iteration low-rank adaptation (LoRA) and extensive uptraining~\cite{bae2025relaxed}. Retrofitted recurrence converts 1B models with a curriculum and about 50B tokens, after which recurrence helps on mathematics~\cite{mcleish2025teaching}. Encode--Think--Decode loops a few mid-depth layers while re-running the mid-training of a base model (50B tokens for OLMo-2 1B) and reports large reasoning gains~\cite{koishekenov2025etd}. Closing the Loop \citep{closingloop2026} converts Qwen3-1.7B with 60--84B tokens. The converted model in this case beats identically continued dense training, but it still trails a larger base model~\cite{closingloop2026}.

\para{Loops added cheaply.} Closest to our setting, Chen et~al.~\cite{chen2026training} loop frozen checkpoints without any training and find that the best window sits at 0.45--0.60 of depth across nine architectures. CoLa and PoLar decide, for each input, which layers of a frozen model to skip or repeat~\cite{li2025cola,li2026polar}, and Sun et~al.~\cite{sun2025painters} repeat or reorder its middle layers. Depth up-scaling duplicates layers and then continues training~\cite{kim2023solar,wu2024llamapro}. In the opposite direction, removing many deep layers costs little accuracy~\cite{gromov2024unreasonable,men2024shortgpt}, and early exit and mixture-of-depths skip computation adaptively~\cite{schuster2022confident,elhoushi2024layerskip,raposo2024mod}. These works show that a carefully placed cheap loop can improve on the model it starts from. However, they do not test whether the gain survives two simple baselines, namely the same model trained identically without the loop and other potential uses of the inference compute introduced by the loop.

\para{Matched comparisons.} Parameter count, FLOPs and latency often rank models differently, and weight-shared models are the standard example~\cite{dehghani2022misnomer}. Sharing weights saves parameters but not computation. ALBERT shares all layers~\cite{lan2020albert}, Takase and Kiyono~\cite{takase2023lessons} compare sharing schemes at fixed parameter budgets, and MobileLLM repeats blocks in small on-device models, gaining accuracy at only a small latency cost~\cite{liu2024mobilellm}. Compute-optimal scaling~\cite{kaplan2020scaling,hoffmann2022chinchilla} is the reference for whether compute is better spent on depth or on parameters and data. Most matched comparisons of recurrence train models from scratch. In that setting, looped models trail iso-FLOP dense models in perplexity but reason well~\cite{saunshi2025reasoning}, and recursion underperforms at the smallest scale~\cite{bae2025mor}. One recurrence is worth about half a unique block~\cite{schwethelm2026isodepth}. Loops also pay when per-token FLOPs, parameters and key--value (KV) cache all match, with a narrower hidden dimension and more experts making up the difference~\cite{smelt2026}, and looped mixture-of-experts models beat standard models of equal pretraining cost~\cite{loopies2026}. For loops added after pretraining, TaH2~\cite{tah2026} finds that fixed-depth looped post-training of Qwen3 base models underperforms at matched test-time compute, while adaptive iteration helps. We extend this comparison in three ways. We use identically trained twins, we give budget forcing to both models, and we cover a range of model sizes.

\para{Cheaper uses of test-time compute.} Instead of more parameters, test-time compute can also go into longer reasoning or more samples~\cite{snell2024scaling,wu2025inference}, and coverage from repeated sampling often grows log-linearly with the number of samples~\cite{brown2024monkeys}. Both decoding rules~\cite{muennighoff2025s1} and training~\cite{aggarwal2025l1} can control the length of reasoning. Reasoning models over-think easy problems~\cite{chen2024overthinking}, and repetition is a long-standing failure of greedy decoding~\cite{holtzman2020curious}. A decoding rule that costs nothing per token may therefore fix the same problem that a loop fixes, for example an answer that never finishes.

Together, these works describe three ways to obtain a looped model, at very different training budgets (Table~\ref{tab:budgets}). Loops learned during pretraining, on hundreds of billions to trillions of tokens, or by converting a model on tens of billions of tokens beat identically trained dense models. Loops added cheaply, with little or no training, can improve on their starting model, but only when applied carefully, since naively repeating layers usually hurts~\cite{chen2026training}. Two questions remain open. First, \emph{does a cheap loop beat both the same model trained identically and the best unlooped use of its compute?} Second, \emph{does the answer to this question vary with model size?}

% =====================================================================================================
\section{Two ways to test a loop}
\label{sec:protocol}

\para{Cost.} A loop re-runs a contiguous span of $m$ of a model's $N$ layers for every token. Each token therefore takes $N{+}m$ layer passes, and its inference cost grows by a factor
\begin{equation}
c=\frac{N+m}{N}.
\label{eq:cost}
\end{equation}
The loops we study have $c=1.25$--$1.29$. Since decoding is memory-bound, each generated token slows down by $\sim c$. Each extra pass also stores its own KV cache.

\para{Two tests.} Let $L$ be a looped model and $U$ its unlooped \emph{twin}, i.e.\ the same model trained on the same batches, in the same order and for the same steps, but without the loop. Let $\mathrm{Acc}(e)$ denote the benchmark accuracy of an evaluation $e$, and $\mathrm{cost}(e)$ its inference FLOPs, computed as (prompt $+$ generated tokens) $\times$ per-token FLOPs. We define the two tests as
\begin{align}
\Delta_{\text{param}} &= \mathrm{Acc}(L)-\mathrm{Acc}(U), \label{eq:param}\\
\Delta_{\text{FLOP}} &= \mathrm{Acc}(L)-\max_{e\in\mathcal{U},\ \mathrm{cost}(e)\le\mathrm{cost}(L)}\mathrm{Acc}(e), \label{eq:flop}
\end{align}
where $\mathcal{U}$ contains the twin at token budgets of 0.5, 1, 1.5 and 2$\times$ the default and, where available, a larger unlooped model. A loop pays only if both differences are positive. The $\Delta_{\text{param}}$ test ignores the loop's compute and asks whether the loop helps at all. Hence a loop with $\Delta_{\text{param}} \le 0$ never reaches the next test. The $\Delta_{\text{FLOP}}$ test picks the best possible comparable use of compute, which favours the unlooped model.

\para{A simple free baseline.} Small reasoning models often never finish their answer, because their reasoning repeats until the token budget runs out. A loop may help the model finish, but so may a decoding rule. \emph{Budget forcing}~\cite{muennighoff2025s1} ends the reasoning once a token limit is reached. We set this limit 512 tokens before the budget, so the model always writes an answer, and it costs nothing extra per token. Whenever we use budget forcing, we give it to both models for fair comparison.

\para{Placing the loop.}\label{sec:place} We want to test a good loop rather than an arbitrary one, without using any evaluation benchmarks to choose it. For the training-free, conversion and post-training loops, we therefore search over candidate spans and score each one on held-out reasoning text. The score is perplexity after temperature scaling~\cite{guo2017calibration}, which divides each candidate's logits by the one temperature that minimises its loss on a validation split. We use this temperature only to score candidates, and never for decoding. The correction is necessary because post-trained models are over-confident, and many loops lower plain perplexity only by flattening the output distribution (Appendix~\ref{sec:search}). The searched Qwen3.5 span also loops one multilayer-perceptron (MLP) block, which raises its cost to $c=1.29$ instead of $1.25$. The ARD, continued-pretraining and from-scratch loops instead use a fixed mid-depth span that we choose before training.

\para{Models and recipes.} We test four kinds of loop, detailed in Appendix~\ref{app:detail}.
\begin{itemize}[leftmargin=*,itemsep=0pt,topsep=2pt]
\item A training-free loop in \emph{post-trained} Qwen3-8B~\cite{yang2025qwen3,chen2026training} reasoning model and the same loop trained into the model with a 33M-token LoRA~\cite{hu2022lora} conversion.
\item Loops that Qwen3.5 \emph{base} models~\cite{qwen35} learn during post-training, with three recipes. The \emph{short} recipe fine-tunes on 100M tokens of reasoning traces (0.8B, 4B, 9B). The second recipe applies LoRA fine-tuning followed by reinforcement learning (RL) at 4B and 9B. The \emph{long} recipe fine-tunes on 400--464M tokens (4B, 9B).
\item Anchored recurrent depth (ARD), a recipe that trains the 4B model to use its extra passes, while a KL anchor keeps its one-pass output close to the original model's (\S\ref{sec:timing}).
\item Continued pretraining of Qwen3.5-0.8B and 4B through the loop for up to 5.2B tokens (\S\ref{sec:budget}).
\end{itemize}
Our main benchmarks are GSM8K~\cite{cobbe2021gsm8k} and MATH-500~\cite{hendrycks2021math,lightman2024verify}. For the long fine-tunes and ARD, we also use AIME 2024/2025 and GPQA-Diamond~\cite{rein2024gpqa}, which none of our models comes close to saturating. We usually report \emph{macro accuracy}, i.e.\ the unweighted mean of the accuracies on the benchmarks in question, so that each benchmark counts equally regardless of its size. We call a difference significant when a paired bootstrap over test items gives $p<0.05$. Appendix~\ref{app:detail3} reports evaluated seeds and intervals.

% =====================================================================================================
\section{Experiments and Results}
\label{sec:post}

\subsection{Only the smallest model gains at equal parameters}
\label{sec:exploit}\label{sec:conv}

We first apply the equal-parameter test to every learned loop (Table~\ref{tab:summary}). Only the 0.8B loop helps. After the same short fine-tune, it leads its twin by 4.1 points (48.3\% vs 44.2\%, $p<0.001$), and it is ahead in all four training seeds. Part of this gain comes from ending answers that would otherwise run until the token budget runs out, but not all of it. On the GSM8K problems that both models finish, the looped model also answers correctly more often (Appendix~\ref{app:detail3}). At 4B and 9B, none of the three post-training recipes changes accuracy significantly, and ARD even lowers it (\S\ref{sec:timing}). The training-free loop in Qwen3-8B brings no significant gain on GSM8K or MATH-500, and training the same loop into the model does not help either (Appendix~\ref{appx:conv}).

\begin{table}[t]
\centering
\caption{\textbf{Equal parameters (Eq.~\ref{eq:param}): only the 0.8B loop helps.} Macro accuracy (\%), twin $\to$ looped, with paired-bootstrap $p$. Green and red bold entries mark a significantly better or worse looped model. The seeds column counts training seeds and, in brackets, those favouring the loop. The benchmarks column lists the benchmarks in the macro average, where MATH is MATH-500, AIME24/25 is AIME 2024/2025 and GPQA is GPQA-Diamond. FT is fine-tuning, CPT continued pretraining and SFT supervised fine-tuning. $^\ddagger$ marks the training-free loop (McNemar test~\cite{mcnemar1947note}) and $^\triangle$ results with fewer than three seeds. Shaded rows mark the key results.}
\label{tab:summary}
\small
\setlength{\tabcolsep}{3pt}
\resizebox{\linewidth}{!}{%
\begin{tabular}{llcccc}
\toprule
Model & Loop learned by & Layer passes (cost) & Seeds & Benchmarks & Twin $\to$ looped \\
\midrule
Qwen3-8B$^\ddagger$ & none (training-free) & 36 $\to$ 45 (1.25$\times$) & -- & GSM8K & 89.4 $\to$ 91.4 {\scriptsize ($p=0.06$)} \\
 &  &  &  & MATH & 76.0 $\to$ 76.6 {\scriptsize ($p=0.74$)} \\
\rowcolor{keyrow} Qwen3.5-0.8B & short full FT & 24 $\to$ 30 (1.25$\times$) & 4 (4/4) & GSM8K, MATH & \textcolor{pos}{\textbf{44.2 $\to$ 48.3 {\scriptsize ($p<0.001$)}}} \\
Qwen3.5-0.8B & 5.2B CPT, then short FT & 24 $\to$ 30 (1.25$\times$) & 2 (2/2) & GSM8K, MATH & \textcolor{pos}{\textbf{31.6 $\to$ 37.6 {\scriptsize ($p<0.001$)}}}$^\triangle$ \\
Qwen3.5-4B & short full FT & 32 $\to$ 40 (1.25$\times$) & 2 (1/2) & GSM8K, MATH & 82.7 $\to$ 82.3 {\scriptsize ($p=0.55$)} \\
Qwen3.5-9B & short full FT & 32 $\to$ 40 (1.25$\times$) & 2 (0/2) & GSM8K, MATH & 86.5 $\to$ 85.9 {\scriptsize ($p=0.21$)} \\
Qwen3.5-4B & LoRA SFT + RL & 32 $\to$ 41 (1.29$\times$) & 2 (1/2) & GSM8K, MATH, AIME24 & 65.1 $\to$ 66.3 {\scriptsize ($p=0.42$)} \\
Qwen3.5-9B & LoRA SFT + RL & 32 $\to$ 41 (1.29$\times$) & 1 (1/1) & GSM8K, MATH, AIME24 & 71.9 $\to$ 72.5 {\scriptsize ($p=0.86$)} \\
Qwen3.5-4B & long full FT & 32 $\to$ 41 (1.29$\times$) & 4 (2/4) & GSM8K, MATH, AIME24/25, GPQA & 59.7 $\to$ 59.7 {\scriptsize ($p=0.98$)} \\
Qwen3.5-9B & long full FT & 32 $\to$ 41 (1.29$\times$) & 1 (0/1) & GSM8K, MATH, AIME24/25, GPQA & 67.6 $\to$ 66.3 {\scriptsize ($p=0.16$)} \\
Qwen3.5-4B & ARD, before post-training & 32 $\to$ 40 (1.25$\times$) & 2 (1/2) & GSM8K, MATH, AIME24/25, GPQA & 59.2 $\to$ 58.7 {\scriptsize ($p=0.50$)} \\
\rowcolor{keyrow} Qwen3.5-4B & ARD, after post-training & 32 $\to$ 40 (1.25$\times$) & 2 (0/2) & GSM8K, MATH, AIME24/25, GPQA & \textcolor{neg}{\textbf{60.3 $\to$ 51.0 {\scriptsize ($p<0.001$)}}}$^\triangle$ \\
\bottomrule
\end{tabular}}
\end{table}

\subsection{At smaller scales, the gain is worth its compute}

Since the 0.8B loop passes the first test, we next ask whether its compute would be better spent on longer answers. Figure~\ref{fig:frontier} plots accuracy against inference cost over four token budgets. At 0.8B, every looped curve lies above the matching twin curve. At 4B and 9B, the curves overlap, and no looped curve lies significantly above its twin. For the training-free loop, a larger model is a better use of the compute. Qwen3-14B takes about 1.3$\times$ the time of looped Qwen3-8B and scores significantly higher on MATH-500 (Table~\ref{tab:eqc}).

\begin{table}[t]
\centering
\caption{\textbf{Equal FLOPs (Eq.~\ref{eq:flop}): the loop wins only at 0.8B.} Macro accuracy (\%), unlooped comparator $\to$ looped, with notation as in Table~\ref{tab:summary}. $^\S$The best cheaper evaluation is the most accurate unlooped one with no greater FLOPs. The best of 12 spans three fine-tuning lengths and four budgets. BF denotes budget forcing on both models. The same-budget comparator is the twin at the loop's own token budget, and Holm $p$ is corrected over the token budgets. Green and red bold entries mark a significant gain or loss for the loop. Shaded rows mark the key results.}
\label{tab:eqc}
\small
\setlength{\tabcolsep}{3pt}
\resizebox{\linewidth}{!}{%
\begin{tabular}{lllcccc}
\toprule
Model & Loop learned by & Unlooped comparator & Comparator cost & Seeds & Benchmarks & Unlooped $\to$ looped \\
\midrule
Qwen3.5-0.8B & short full FT & best cheaper$^\S$ & $\le$ loop & 4 (4/4) & GSM8K, MATH & \textcolor{pos}{\textbf{45.1 $\to$ 48.0 {\scriptsize ($p<0.001$)}}} \\
 & \quad both + BF & best cheaper, BF$^\S$ & $\le$ loop & 4 (4/4) & GSM8K, MATH & \textcolor{pos}{\textbf{54.8 $\to$ 57.3 {\scriptsize ($p<0.001$)}}} \\
\rowcolor{keyrow}  & \quad both 3$\times$ FT + BF & best of 12, BF$^\S$ & $\le$ loop & 2 (2/2) & GSM8K, MATH & \textcolor{pos}{\textbf{60.2 $\to$ 64.1 {\scriptsize ($p<0.001$)}}}$^\triangle$ \\
Qwen3.5-0.8B & 5.2B CPT, then short FT & best cheaper$^\S$ & $\le$ loop & 2 (2/2) & GSM8K, MATH & \textcolor{pos}{\textbf{32.8 $\to$ 37.2 {\scriptsize ($p<0.001$)}}}$^\triangle$ \\
 & \quad both + BF & best cheaper, BF$^\S$ & $\le$ loop & 2 (2/2) & GSM8K, MATH & \textcolor{pos}{\textbf{40.4 $\to$ 46.4 {\scriptsize ($p<0.001$)}}}$^\triangle$ \\
\midrule
Qwen3.5-4B & short full FT & best cheaper$^\S$ & $\le$ loop & 2 (1/2) & GSM8K, MATH & 83.4 $\to$ 82.5 {\scriptsize ($p=0.41$)} \\
 & \quad both + BF & best cheaper, BF$^\S$ & $\le$ loop & 2 (0/2) & GSM8K, MATH & 88.2 $\to$ 87.7 {\scriptsize ($p=0.33$)} \\
Qwen3.5-9B & short full FT & best cheaper$^\S$ & $\le$ loop & 2 (0/2) & GSM8K, MATH & 86.4 $\to$ 85.9 {\scriptsize ($p=0.34$)} \\
 & \quad both + BF & best cheaper, BF$^\S$ & $\le$ loop & 2 (1/2) & GSM8K, MATH & 90.2 $\to$ 90.2 {\scriptsize ($p=1.00$)} \\
\midrule
\rowcolor{keyrow} Qwen3.5-4B & long full FT & best cheaper$^\S$ & $\le$ loop & 2 & GSM8K, MATH & \textcolor{neg}{\textbf{86.4 $\to$ 85.1 {\scriptsize (Holm $p=0.03$)}}}$^\triangle$ \\
Qwen3.5-4B & long full FT, both + BF & same budget, BF & $\le$ loop & 2 (1/2) & GSM8K, MATH & 88.9 $\to$ 89.2 {\scriptsize ($p=0.47$)} \\
Qwen3.5-9B & long full FT, both + BF & same budget, BF & $\le$ loop & 1 (0/1) & GSM8K, MATH & 91.4 $\to$ 91.2 {\scriptsize ($p=0.78$)} \\
\midrule
\multicolumn{7}{l}{\textit{The comparator spends more time than the loop}} \\
Qwen3-8B$^\ddagger$ & none (training-free) & Qwen3-14B & higher (1.28$\times$) & -- & GSM8K & 90.2 $\to$ 91.4 {\scriptsize ($p=0.46$)} \\
\rowcolor{keyrow}  &  &  & higher (1.34$\times$) & -- & MATH & \textcolor{neg}{\textbf{80.2 $\to$ 76.6 {\scriptsize ($p=0.01$)}}} \\
Qwen3.5-9B & long full FT & 1.25$\times$ generated tokens & higher (1.16$\times$) & 1 (0/1) & GSM8K, MATH, AIME24/25, GPQA & \textcolor{neg}{\textbf{70.5 $\to$ 66.3 {\scriptsize ($p<0.001$)}}}$^\triangle$ \\
\bottomrule
\end{tabular}}
\end{table}

\begin{figure}[t]
\centering
\includegraphics[width=\linewidth]{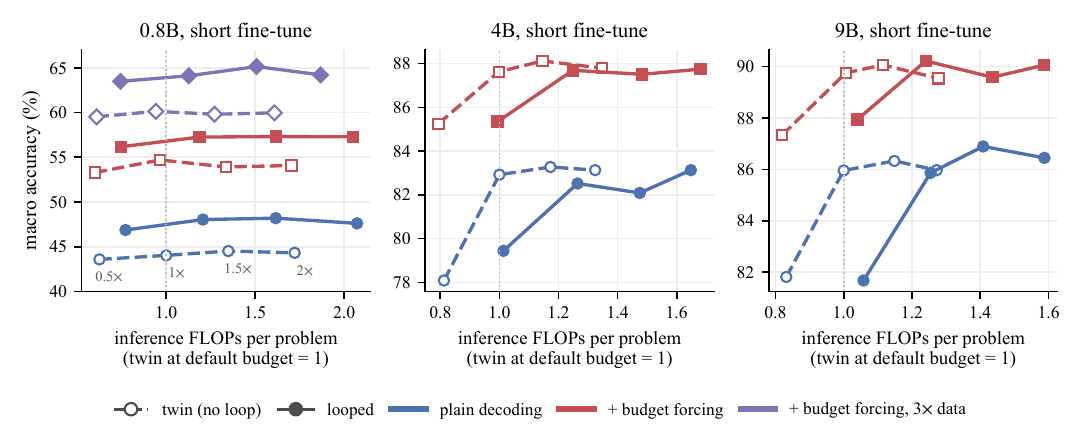}
\caption{\textbf{At equal FLOPs, the loop is ahead only at 0.8B.} The y-axis is macro accuracy on GSM8K and MATH-500. The x-axis is the inference compute spent per problem, i.e.\ (prompt $+$ generated tokens) $\times$ per-token FLOPs, divided by the twin's compute at the default token budget, so the twin at its default budget sits at 1 (dotted line). Each line connects four token budgets, 0.5, 1, 1.5 and 2$\times$ the default (labelled at 0.8B). Each colour is one setting that both models receive, with the twin dashed and the looped model solid. The loop pays at equal FLOPs where the solid line lies above the dashed line of the same colour at the same $x$. Only the 0.8B panel includes the setting with three times the fine-tuning data.}
\label{fig:frontier}
\end{figure}

\begin{table}[t]
\centering
\caption{\textbf{Budget forcing and more fine-tuning at 0.8B} (short fine-tune, default budget, seeds 0--1). FT counts fine-tuning tokens. The repetition stop is a second free decoding rule that ends the reasoning once the output repeats the same stretch of 4--256 tokens three times. The finished column gives the GSM8K and MATH-500 answers completed within the budget, and cost gives FLOPs relative to the plain twin. Adding seeds 2--3 to the settings that have them changes no accuracy by more than 1.1 points, so the results do not depend on the choice of seeds. Shaded rows mark the key comparisons, and bold numbers are the macro accuracies that the text compares. Budget forcing alone lifts the twin above the plain looped model, and with both fixes on both models the loop still leads while both finish almost every GSM8K answer.}
\label{tab:fix}
\small
\setlength{\tabcolsep}{4pt}
\begin{tabular}{lccccc}
\toprule
Setting & GSM8K & MATH-500 & Macro & Finished (\%) & Cost ($\times$) \\
\midrule
Unlooped & 54.7 & 33.0 & 43.9 & 72 / 44 & 1.00 \\
\rowcolor{keyrow} Looped & 61.2 & 35.1 & \textbf{48.2} & 76 / 44 & 1.20 \\
\midrule
Unlooped + repetition stop & 59.7 & 40.4 & 50.0 & 86 / 67 & 0.80 \\
\rowcolor{keyrow} Unlooped + budget forcing & 63.8 & 44.6 & \textbf{54.2} & 97 / 73 & 0.96 \\
Looped + budget forcing & 68.5 & 44.6 & 56.5 & 97 / 68 & 1.18 \\
\midrule
Unlooped, 1.25$\times$ FT + budget forcing & 64.1 & 46.2 & 55.2 & 97 / 71 & 0.96 \\
\rowcolor{keyrow} Unlooped, 3$\times$ FT + budget forcing & 68.4 & 51.9 & \textbf{60.1} & 97 / 76 & 0.95 \\
\rowcolor{keyrow} Looped, 3$\times$ FT + budget forcing & 74.0 & 54.2 & \textbf{64.1} & 98 / 76 & 1.13 \\
\bottomrule
\end{tabular}
\end{table}

A harder question is whether the 0.8B gain is more than a fix for unfinished answers. Budget forcing finishes answers at no cost. On its own, it lifts the unlooped 0.8B model from 44\% to 54\%, which is above the looped model without budget forcing (48\%, Table~\ref{tab:fix}). Giving budget forcing to the unlooped model alone would therefore hide the loop's gain, while giving it to both models measures that gain. With budget forcing on both models, the loop still leads at equal FLOPs (57.3\% vs 54.8\%). When both models also receive three times the fine-tuning data, the loop leads by four points (Table~\ref{tab:eqc}). Both models now finish almost every GSM8K answer (Table~\ref{tab:fix}), so the remaining gain does not come from finishing. Hence a fair comparison must match decoding and training on both sides. Otherwise it credits one model with a fix that the other could also use.

\subsection{Larger models do not fail because of a ceiling}
\label{sec:headroom}

A null result at 4B and 9B has two possible explanations. Either the loop adds nothing that a larger model can use, or the benchmarks are too easy to show what it adds. To tell them apart, we ask two separate questions. First, does a larger model become more accurate than its twin when we add a loop, ignoring compute? Secondly, does it improve \emph{in proportion to} the compute that the loop adds? The second question matters only if the answer to the first is yes, and the first test already says no. Where a loop even lowers accuracy, as ARD does after post-training (\S\ref{sec:timing}), its compute overhead does not matter, and the loop should simply not be added.

\begin{table}[t]
\centering
\caption{\textbf{The benchmarks leave room for a gain} (long fine-tune, accuracy in \%). The size gap is the 9B model minus the 4B model, and the loop columns give $\Delta_{\text{param}}$. The 4B results average four seeds, and the 9B results come from one. Shaded rows mark the key results. Bold marks the two largest size gaps, on AIME, where the 4B loop still changes accuracy by less than one point.}
\label{tab:headroom}
\small
\setlength{\tabcolsep}{5pt}
\begin{tabular}{lccccc}
\toprule
Benchmark & 4B & 9B & Size gap & Loop at 4B & Loop at 9B \\
\midrule
GSM8K & 90.0 & 93.7 & \pls 3.7 & \phantom{\mns}0.0 & \mns 1.1 \\
MATH-500 & 81.5 & 86.0 & \pls 4.5 & \mns 0.5 & \mns 0.8 \\
\rowcolor{keyrow} AIME 2024 & 49.3 & 61.7 & \textbf{\pls 12.4} & \pls 0.4 & \mns 2.1 \\
\rowcolor{keyrow} AIME 2025 & 41.1 & 53.8 & \textbf{\pls 12.7} & \pls 0.5 & \mns 4.6 \\
GPQA-Diamond & 36.5 & 42.9 & \pls 6.4 & \mns 0.3 & \pls 2.0 \\
\midrule
Macro & 59.7 & 67.6 & \pls 7.9 & \phantom{\mns}0.0 & \mns 1.3 \\
\bottomrule
\end{tabular}
\end{table}

The benchmarks leave room for a gain. GSM8K is close to its ceiling, but AIME and GPQA-Diamond are not. The unlooped 4B model solves fewer than half of their problems, and the 9B model scores 6--13 points higher (Table~\ref{tab:headroom}). On the same benchmarks, the 4B loop changes accuracy by less than one point. A rough estimate shows how large a gain the loop would need. If accuracy grows log-linearly with per-token compute between 4B and 9B, and per-token FLOPs scale with parameter count, then a loop that multiplies the per-token cost by $c=(N{+}m)/N$ (Eq.~\ref{eq:cost}) is worth
\begin{equation}
\Delta^{\star}\approx\frac{\log c}{\log\left(P_{9\text{B}}/P_{4\text{B}}\right)}\left(\mathrm{Acc}_{9\text{B}}-\mathrm{Acc}_{4\text{B}}\right)=\frac{\log1.29}{\log2.25}\times7.9\approx2.5\ \text{points},
\label{eq:yardstick}
\end{equation}
where $P$ denotes parameter count. For the long fine-tune, the loop takes 41 instead of 32 layer passes, so $c=1.29$. The 9B model has about 2.25 times the parameters of the 4B model, and it scores 7.9 points higher on the macro average (Table~\ref{tab:headroom}). The 95\% interval of the 4B loop's effect, $\pm1.1$ points, excludes this value. Equation~\eqref{eq:yardstick} is a back-of-the-envelope estimate and not a scaling law. It nevertheless shows that the benchmarks can reveal a gain of the size that the loop needs.

Room to improve does not explain the 0.8B result either. The 0.8B model has the most room, and part of its gain comes from finishing answers. However, the 4B model also has ample room on AIME and GPQA-Diamond, and it does not gain. Why small models benefit from added depth therefore remains an interesting open question.

\subsection{Anchoring the loop to the base model does not help}
\label{sec:timing}

Plain looped training may fail because the model never learns to use its second pass. Anchored recurrent depth (ARD) targets exactly this failure. It trains one LoRA adapter, shared by all passes, and adds three terms to the task loss (Appendix~\ref{app:detail}).
\begin{itemize}[leftmargin=*,itemsep=0pt,topsep=2pt]
\item An \emph{anchor}, a Kullback--Leibler (KL) divergence, keeps the one-pass model close to the original, so looping does not erase what the model knows.
\item A \emph{consistency} term pulls the predictions of shallower passes toward those of the deepest pass.
\item A \emph{depth reward} is positive when two passes make the correct answer more likely than one pass.
\end{itemize}
We add ARD to Qwen3.5-4B either before post-training, as 50M tokens of continued pretraining, or after post-training, on 50M tokens of reasoning traces. Each setting has a control with the same adapter, data and steps, but without the loop. After post-training, ARD lowers accuracy by nine points, and accuracy falls further with every extra pass. Hence the model never learns to use the loop. Before post-training, ARD is close to neutral (Table~\ref{tab:summary}). The timing of the loop therefore matters, but in neither position does the loop beat its twin.

\subsection{When is a loop worth its cost?}
\label{sec:serving}

On one H100, the loop slows each generated token by 1.2--1.3$\times$ and cuts throughput by about a fifth (Appendix~\ref{app:serving}). Whether this cost is worth paying depends on what else the budget allows. A looped 0.8B model keeps the disk size, memory footprint and training cost of a 0.8B model. An unlooped 4B model scores much higher than the looped 0.8B model (83\% against 48\% after the same short fine-tune), and it decodes only 1.4$\times$ slower per token. However, it stores five times as many parameters, needs about five times the memory, and costs about five times the compute per training token and, being larger, needs more training tokens~\cite{hoffmann2022chinchilla}. A loop is therefore only useful under a fixed parameter budget, for example for an on-device model or with a limited training budget. When the budget can grow, a larger model is the better choice.

% =====================================================================================================
\section{More training through the loop}
\label{sec:budget}

\begin{figure}[t]
\centering
\includegraphics[width=\linewidth]{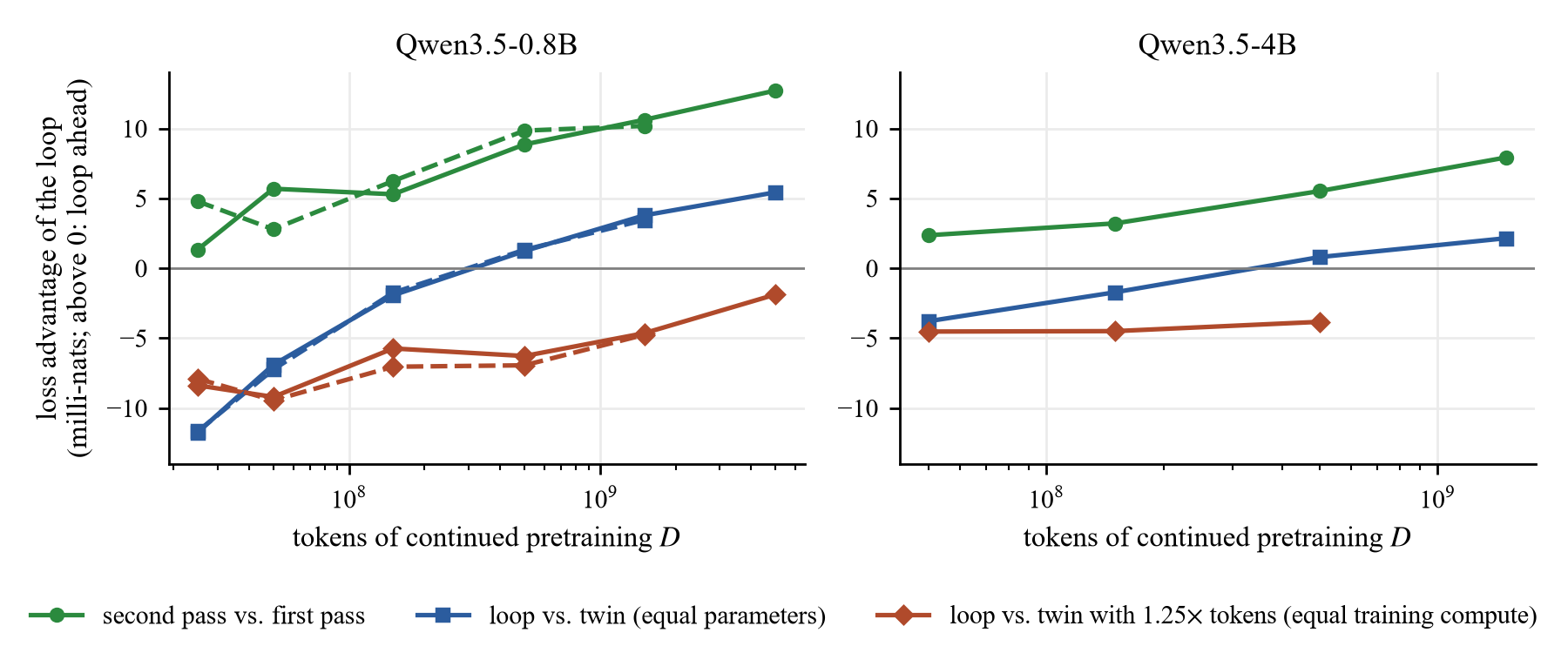}
\caption{\textbf{Continued pretraining through the loop.} Each curve is the held-out loss of a comparison model minus that of the looped model, in milli-nats, so values above zero mean that the loop is ahead. Green compares the looped model's second pass with its own first pass, blue compares the looped model with its twin (equal parameters), and red compares it with a twin trained on 1.25$\times$ the tokens (equal training compute). Dashed lines show the second 0.8B seed. The 4B panel has one seed, and its red line stops at 0.5B tokens because the twin run does not extend to 1.25$\times$ the later checkpoints.}
\label{fig:sweep}
\end{figure}

\begin{table}[!htbp]
\centering
\caption{\textbf{Loops learned from scratch} (113M parameters, FineWeb-Edu). Training FLOPs and inference cost are relative to the dense model, loss is in nats per token, and zero-shot accuracy is in \%. Test depth $K$ is the number of passes through the looped core at evaluation. The conversion row is the dense model trained on 1.2$\times$ the tokens, with the loop added for the extra tokens. Shaded rows mark the key results. The bold losses compare the looped model with a dense model given the same training FLOPs, which reaches the lower loss.}
\label{tab:pretrain}
\small
\setlength{\tabcolsep}{3pt}
\resizebox{\linewidth}{!}{%
\begin{tabular}{lcccccccc}
\toprule
Model & Tokens (B) & Train FLOPs & Inference & Loss & HellaSwag & ARC-E & ARC-C & LAMBADA \\
\midrule
Dense, pretrained without loop & 1.50 & 1.00 & 1.00 & 3.308 & 28.3 & 42.9 & 22.9 & 18.9 \\
\rowcolor{keyrow} Dense, FLOP-matched to looped pretraining & 1.83 & 1.22 & 1.00 & \textbf{3.250} & 28.5 & 43.5 & 24.3 & 20.4 \\
\rowcolor{keyrow} Looped during pretraining & 1.50 & 1.22 & 1.22 & \textbf{3.289} & 28.4 & 43.5 & 23.8 & 19.2 \\
Dense, 16 layers & 1.50 & 1.22 & 1.22 & 3.268 & 28.4 & 43.6 & 24.5 & 19.2 \\
Dense + best searched loop (no training) & 1.50 & 1.00 & 1.28 & 3.306 & 28.2 & 43.0 & 22.8 & 19.1 \\
Dense, 1.2$\times$ tokens, no loop & 1.80 & 1.20 & 1.00 & 3.278 & 28.3 & 42.8 & 23.7 & 19.4 \\
Dense, 1.2$\times$ tokens through the loop (conversion) & 1.80 & 1.24 & 1.22 & 3.273 & 28.4 & 42.8 & 23.9 & 19.9 \\
Random depth, test depth $K{=}2$ & 1.50 & 1.22 & 1.22 & 3.318 & 28.2 & 42.0 & 23.0 & 17.3 \\
Input-injected recurrence, test depth $K{=}4$ & 1.50 & 1.47 & 1.71 & 3.389 & 27.9 & 42.2 & 23.0 & 15.8 \\
\bottomrule
\end{tabular}}
\end{table}

Successful loops learn from tens of billions of tokens (Table~\ref{tab:budgets}), whereas our post-training loops see at most half a billion. We therefore ask whether more training changes the picture, first by continuing to pretrain through the loop and then by learning the loop from scratch.

\para{Continued pretraining.} We continue pretraining the 0.8B and 4B base models through the loop for up to 5.2B tokens (Figure~\ref{fig:sweep}). The blue curve, the looped model's loss advantage over its twin, crosses zero at about 0.5B tokens, so from then on the looped model reaches a slightly lower held-out loss than its twin. Closing the Loop~\cite{closingloop2026} also finds a converted looped model ahead of an identically continued dense model, but it reports this only at the end of 60B or 84B tokens of conversion. In our runs, the looped model is already ahead after about 0.5B tokens, a budget two orders of magnitude smaller. After the same short fine-tune, the looped model also scores higher than its twin (Appendix~\ref{app:sec4}). However, the red curve stays below zero, so within the 5.2B tokens we train, the looped model never reaches the loss of a twin given the same training compute. Moreover, continued pretraining on our web-and-math data leaves every model below a plain fine-tune of the base model.

\para{Loops learned from scratch.} We also train a 113M-parameter model from scratch, with and without the loop, alongside dense and recurrent controls (Table~\ref{tab:pretrain}, details in Appendix~\ref{app:pretrainfull}). The looped model reaches a held-out loss of 3.289 nats against 3.308 for its twin, but a dense model given the same training FLOPs reaches 3.250, and a 16-layer dense model reaches 3.268. Adding the same loop after pretraining, without training, barely changes the loss, and neither random depth nor an input-injected core~\cite{geiping2025scaling} reaches the FLOP-matched dense model. More training therefore moves a loop past its twin, but within the budgets we can afford, not past a model given equal training compute.

% =====================================================================================================
\section{Conclusion}
\label{sec:conclusion}

There is no free loop. Re-applying $m$ of $N$ layers makes every generated token cost $(N{+}m)/N$ times as much, and this cost buys accuracy only for the smallest model. So far, useful loops have been learned only with large budgets, either during pretraining or by converting a model on tens of billions of tokens. At 4B, even our longer fine-tune did not change this. A budget close to that of pretraining might, but at that point one could simply pretrain a looped model. Three observations explain why comparing a loop only with its starting model can make it look useful when it is not. First, comparing a looped model with the model it started from credits the loop with its extra training and its extra compute. Our two baselines, an identically trained twin and the best unlooped use of the same compute, remove both confounding factors. Second, a decoding rule that costs nothing does much of what a loop does for a small model, namely finishing its answers. Hence both models must receive the rule. Third, a null result at larger scale is informative only if the benchmarks can reveal a gain of the required size. Ours can, as the gap between 4B and 9B shows. Claims about recurrence should therefore state the model size and report both tests, with decoding and training matched on both sides.

Our study has one main limitation. We test fixed-depth loops that we add with modest budgets and a few recipes, so our conclusions may not hold for other forms of recurrence. 
%In future work, we plan to apply our two tests to adaptive-depth recurrence~\cite{tah2026} and to recurrence learned at pretraining scale~\cite{geiping2025scaling,zhu2025ouro}, which may behave differently at 4B and above. 
In future work, we want to understand why the 0.8B model benefits from the loop while the 4B model, with ample room on hard benchmarks, does not.

\newpage
\appendix

% =====================================================================================================
\section{Setup details}
\label{app:detail}

Appendix~\ref{app:detail} describes the setup. Appendix~\ref{app:detail3} gives the equal-parameter and equal-FLOP results behind Section~\ref{sec:post}, and Appendix~\ref{app:sec4} the results behind Section~\ref{sec:budget}. Appendix~\ref{app:released} covers training-free loops and the Qwen3-8B conversion, and Appendix~\ref{sec:search} the search objective that places the loop.
Table~\ref{tab:notation} lists the notation used throughout.

\begin{table}[htbp]
\centering
\caption{\textbf{Notation.}}
\label{tab:notation}
\small
\begin{tabular}{ll}
\toprule
Symbol & Meaning \\
\midrule
$N$, $m$ & layers of the unlooped model, layers that the loop re-runs \\
$c=(N{+}m)/N$ & inference cost of the looped model relative to its twin (Eq.~\ref{eq:cost}) \\
$L$, $U$ & looped model, its unlooped twin \\
$\mathrm{Acc}(e)$, $\mathrm{cost}(e)$ & accuracy and inference FLOPs of an evaluation $e$ \\
$\mathcal{U}$ & unlooped evaluations available to the equal-FLOP test \\
$\Delta_{\text{param}}$, $\Delta_{\text{FLOP}}$ & gain at equal parameters and at equal FLOPs (Eqs.~\ref{eq:param}--\ref{eq:flop}) \\
$P$ & parameter count (Eq.~\ref{eq:yardstick}) \\
$K$ & number of passes through a looped block \\
$g$, $F$, $\beta$ & looped block, one damped step, anchor of the damped loop (Eq.~\ref{eq:damped}) \\
$T$, $T^*$, $\pplT$ & temperature, fitted temperature, perplexity at the fitted temperature \\
$D$, $r(D)$ & training tokens, token exchange rate \\
\bottomrule
\end{tabular}
\end{table}

% -----------------------------------------------------------------------------------------------------
\subsection{The loop and where it goes}

\para{Two loop operators.} A loop passes its input through the same block $g$ a total of $K$ times. The block is an attention sub-block, an MLP sub-block, one layer, or a window of consecutive layers. Training-free and converted Qwen3 loops use the damped update of Chen et~al.~\cite{chen2026training}. It takes $K$ damped Euler steps, anchored to the trained one-step output,
\begin{equation}
F(x)=x+\tfrac1K\big(g(x)-x\big),\qquad g^{(K)}(x)=\beta\,g(x)+(1-\beta)\,F^{K}(x).
\label{eq:damped}
\end{equation}
Here $g^{(K)}$ names the looped block and is not a power of $g$, and for $K=1$ it is the original layer. The anchor $\beta$ keeps the output close to the input that later layers expect. A damped loop is thus a finer discretization of the same residual update~\cite{lu2019understanding,li2022ode}, not added depth in the sense of universal transformers~\cite{dehghani2019universal}. All loops trained on Qwen3.5 instead use a plain recurrent core, in which the model applies the span again with shared weights.

\para{Placing the loop.} A search that never uses benchmark answers scores candidate spans by calibrated perplexity $\pplT$, the perplexity at the candidate's own fitted temperature (Appendix~\ref{sec:search}). For the Qwen3.5 base models, an exhaustive window search selects an eight-layer window at the start of the network (layers 1--8 for 4B, 0--7 for 9B). The model loops it twice, plus one MLP loop ($K=3$) at its first layer. In these models, windows starting at layers 13--20 raise $\pplT$ above that of the unlooped model. The depth rule of Chen et~al.\ (window centre at 0.45--0.60 of depth) therefore picks a placement that our objective predicts to be harmful, and we add it as a control in the LoRA with reinforcement learning recipe. The other trained loops use a fixed mid-depth span, chosen before training by minimum KL divergence to the unlooped model.

% -----------------------------------------------------------------------------------------------------
\subsection{Matched controls and cost}

Besides its twin, each looped model can have an \emph{equal-compute} control, i.e.\ the twin given the loop's extra training compute (more steps or tokens) or its extra inference compute (a longer token budget, or a larger model). When extra training degrades the twin, the equal-compute comparison uses the best unlooped model at no more than the loop's compute.

We charge the loop its layer passes, $(N{+}m)/N$, which matches the measured slowdown of memory-bound decoding (1.18--1.26$\times$ across our pairs). The searched Qwen3.5 spans also loop one MLP block, so their 32 $\to$ 41 passes cost 1.29$\times$. In Qwen3.5, three of every four layers use linear attention, whose extra passes add recurrent state rather than KV cache.

% -----------------------------------------------------------------------------------------------------
\subsection{How we train each loop}
\label{app:recipes}

This section describes how we train each loop, and Appendix~\ref{app:detail3} reports what each recipe achieves. Table~\ref{tab:budgets} compares our budgets for learning the loop with prior work.

\para{Conversion of Qwen3-8B.} We train LoRA adapters on every linear layer through the looped computation. All iterations share the adapters, and we merge them afterwards, so the model gains no parameters. Following the recurrence curriculum of McLeish et~al.~\cite{mcleish2025teaching}, we introduce the loop gradually, so the probability of applying it rises from 0 to 1 over the first third of training. The loss on reasoning traces is a KL divergence to the unmodified model as teacher (weight 0.9) plus the next-token negative log-likelihood (weight 0.1), which keeps the twin from degrading. The main run takes 2{,}000 steps on 33M tokens of decontaminated data.

\para{Short fine-tune.} We fully fine-tune Qwen3.5-0.8B, 4B and 9B base models on 100M tokens of reasoning traces, with and without the loop.

\para{LoRA fine-tuning with reinforcement learning.} Qwen3.5-4B and 9B get LoRA fine-tuning through the loop on thinking-mode reasoning traces, then group relative policy optimisation (GRPO)~\cite{shao2024deepseekmath} on GSM8K and MATH training problems. The reward is a correct boxed answer after finished thinking. The settings are no loop, the searched span, the span moved 8 layers deeper, the span of the depth rule of Chen et al., and no loop with fine-tuning extended to the looped model's training FLOPs.

\para{Long fine-tune.} We fully fine-tune both Qwen3.5 base models, with and without the searched loop, on 400--464M tokens of decontaminated long reasoning traces (Table~\ref{tab:budgets}) that are never truncated. The traces are verified DeepSeek-R1~\cite{guo2025deepseekr1} solutions from OpenR1-Math and QwQ traces from OpenThoughts3~\cite{guha2026openthoughts}. For 4B, the looped model's throughput within a fixed time budget sets the step count.

\para{Anchored recurrent depth (ARD).} ARD applies a core of 8 of the 32 layers of Qwen3.5-4B $K$ times, with $K$ drawn from a curriculum. It trains one merged LoRA adapter with the task loss plus three terms. A KL anchor ties the model to the frozen depth-1 model, a depth-consistency KL pulls shallower depths toward the deepest one, and a reward favours depth $K$ over depth 1 when it raises the likelihood of the gold answer. \emph{Before} post-training, ARD continues pretraining the base model for 50M tokens on FineWeb-Edu~\cite{penedo2024fineweb} and FineMath~\cite{allal2025smollm2}, followed by the long fine-tune with $K$ drawn from 1--3. \emph{After} post-training, ARD runs for 50M tokens on reasoning traces from the unlooped long fine-tune. Controls use the same adapter, data, steps and anchor without recurrence.

\para{Continued pretraining.} Qwen3.5-0.8B and 4B base models continue pretraining for up to 5.2B tokens on a mix of FineWeb-Edu and FineMath. The looped model applies the loop a random 1--3 times per batch. The learning rate is constant after warm-up, so checkpoints lie on one trajectory. Appendix~\ref{app:pretrainfull} describes the from-scratch models.

% -----------------------------------------------------------------------------------------------------
\subsection{Evaluation}
\label{app:protocol}

\para{Prompts and scoring.} Generative tasks use the chat template with one user turn, which contains the question followed by ``Please reason step by step, and put your final answer within \textbackslash boxed\{\}.'' The answer is the last boxed expression after any thinking block (for GSM8K, the last number if there is no box), compared with the reference after normalising \LaTeX{} and numerically when possible. Non-thinking token budgets are 768 for GSM8K and 2{,}048 for MATH-500. Thinking-mode GSM8K uses 4{,}096. AIME accuracy averages over several sampled answers per problem. Within a seed, loop and control use the same number of sampled answers, except in the decoding-budget control. We score multiple-choice tasks (a fixed 2{,}000-question subset of MMLU-Pro~\cite{wang2024mmlupro}, ARC-Challenge~\cite{clark2018arc} and CommonsenseQA~\cite{talmor2019commonsenseqa}) zero-shot by the highest next-token logit over option letters. This lowers absolute numbers but treats every configuration alike.

\para{Decoding checks.} For exact cached decoding, every looped call site keeps its own KV cache or, in the Gated DeltaNet layers of Qwen3.5, its own recurrent and convolution state. Every evaluation checks cached looped decoding against full recomputation. Decoding is greedy in bfloat16 with left-padded batches. Batch composition changes a small fraction of greedy outputs and moves MATH-500 accuracy by up to 1.2 points across repeated evaluations, so comparisons within a table share one evaluation. Every evaluation must cover all of its items.

\para{Decontamination and data separation.} We remove any training problem that shares a 13-gram of lower-cased words with GSM8K test, MATH-500, AIME 2024/2025 or GPQA-Diamond. This does not detect paraphrases or problems shorter than 13 words. No selection step sees a test question. The search fits temperatures on the validation split of its reference mixture, and accuracy-based selection uses GSM8K and MATH training problems.

% -----------------------------------------------------------------------------------------------------
\subsection{Statistics}

Intervals come from paired bootstraps over test items, conditional on the training seeds. They ignore variation between training runs, so we also report per-seed differences. All $p$-values are two-sided, and $p<0.001$ marks the resampling floor. \emph{Equivalence} means a 90\% interval within $\pm2$ points (two one-sided tests~\cite{schuirmann1987}). Comparisons across token budgets are Holm-corrected~\cite{holm1979simple}.

% -----------------------------------------------------------------------------------------------------
\subsection{Models, data and code}

Training-free experiments use Qwen3 (1.7B to 32B). Training experiments use the Qwen3.5-0.8B, 4B and 9B base models~\cite{qwen35}, which alternate three Gated DeltaNet linear-attention layers with one gated full-attention layer, and a 113M-parameter model trained from scratch. The search text is a domain-balanced mixture of reasoning, code and chat responses from Qwen3-235B~\cite{amteam2025}. We will release code and result files upon acceptance.

% =====================================================================================================
\section{Equal-parameter and equal-FLOP results}
\label{app:detail3}

% -----------------------------------------------------------------------------------------------------
\subsection{Equal-parameter results for each recipe}

\begin{figure}[t]
\centering
\includegraphics[width=0.86\linewidth]{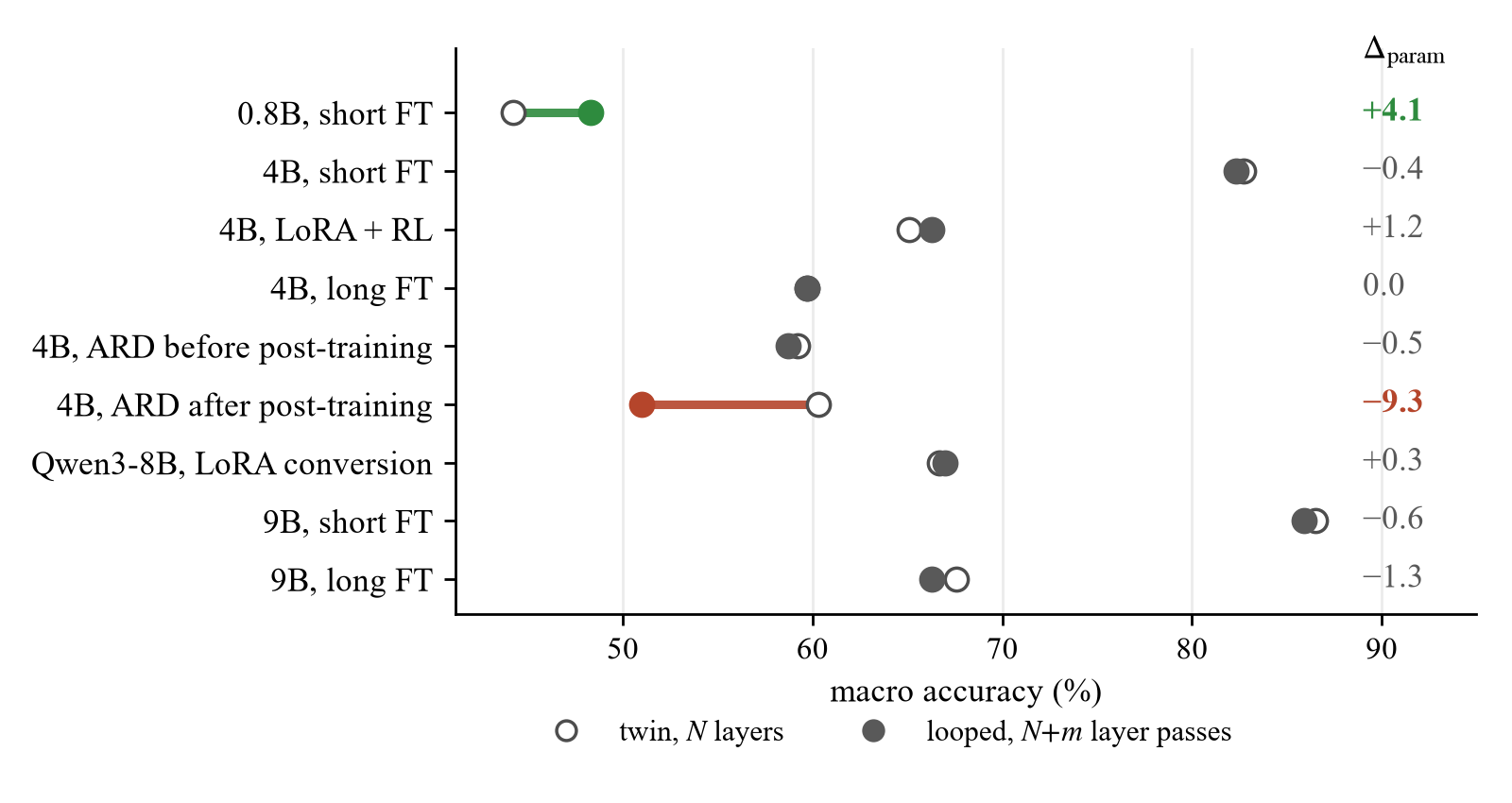}
\caption{\textbf{Only the 0.8B loop helps at equal parameters (Eq.~\ref{eq:param}).} Filled circles show looped models and open circles their twins. The right column gives $\Delta_{\text{param}}$ in points, in green or red when the gain or loss is significant. FT is fine-tuning on 100M (short) or 400--464M (long) tokens, RL is reinforcement learning, and ARD is anchored recurrent depth. Table~\ref{tab:summary} gives the same results with seeds and $p$-values.}
\label{fig:budget}
\end{figure}

Figure~\ref{fig:budget} shows the equal-parameter results of Table~\ref{tab:summary} graphically. The paragraphs below give the details of each recipe.

\para{Short fine-tune results.} The looped 0.8B model scores 48.3\% against 44.2\% for its twin ($p<0.001$). It is ahead in all four seeds and on both GSM8K ($p<0.001$) and MATH-500 ($p=0.005$). Part of the gain is termination. The looped model finishes more GSM8K answers (Table~\ref{tab:fix}), and doubling the budget almost never finishes the rest. Part is not. On GSM8K problems that both models finish, the looped model answers 79.1\% correctly against 75.6\% ($p<0.001$). The loop also beats a twin fine-tuned on 1.25$\times$ the tokens, which matches its training compute (48.6\% vs 45.4\%, $p<0.001$, two seeds). At 4B and 9B, 92--95\% of answers already finish within the budget, so finishing has less to offer, and the loop stays within one point of both controls ($p\ge0.21$).

\para{LoRA fine-tuning with reinforcement learning, results.} After LoRA fine-tuning, the looped 4B model is behind its twin on GSM8K, MATH-500 and AIME 2024. After GRPO, the two are within half a point on GSM8K and MATH-500, and the looped model leads on AIME 2024. Averaged over the three benchmarks, no difference is significant at 4B or 9B (Table~\ref{tab:summary}).

\para{Long fine-tune results.} Across five benchmarks we detect no difference (Table~\ref{tab:maxsft}). The macro average is 59.7\% for both models at 4B ($p=0.98$), with the looped model higher in two of four seeds and the two models equivalent within $\pm2$ points. At 9B it is 66.3\% against 67.6\% ($p=0.16$, one seed). No per-benchmark difference survives Holm correction over the ten contrasts. The looped models take 32--34\% more training time. Against a 4B twin given that extra training (seeds 0--1), the loop scores 60.3\% vs 59.1\% ($p=0.10$).

\begin{table}[t]
\centering
\caption{\textbf{Long fine-tune, per benchmark.} Accuracy (\%), twin $\to$ looped, with the 95\% paired-bootstrap interval and $p$ Holm-adjusted over the ten contrasts. A comparison counts as equivalent when its 90\% interval lies within $\pm2$ points. GPQA-D is GPQA-Diamond. The macro rows, in bold, give the unweighted mean over the five benchmarks. Shaded rows mark the key results.}
\label{tab:maxsft}
\small
\setlength{\tabcolsep}{4pt}
\begin{tabular}{llcccc}
\toprule
Model & Benchmark & Twin$\to$looped & 95\% CI & $p_\text{Holm}$ & Equivalent \\
\midrule
4B (4 seeds) & MATH-500 & 81.5$\to$81.0 & [\mns 1.9, \pls 0.8] & 1.00 & yes \\
4B (4 seeds) & AIME 2024 & 49.3$\to$49.7 & [\mns 1.7, \pls 2.7] & 1.00 & no \\
4B (4 seeds) & AIME 2025 & 41.1$\to$41.6 & [\mns 3.0, \pls 4.2] & 1.00 & no \\
4B (4 seeds) & GPQA-D & 36.5$\to$36.2 & [\mns 3.4, \pls 2.9] & 1.00 & no \\
4B (4 seeds) & GSM8K & 90.0$\to$90.0 & [\mns 0.7, \pls 0.8] & 1.00 & yes \\
\rowcolor{keyrow} 4B (4 seeds) & \textbf{macro} & \textbf{59.7$\to$59.7} & [\mns 1.1, \pls 1.1] & -- & yes \\
\midrule
9B (1 seed) & MATH-500 & 86.0$\to$85.2 & [\mns 3.2, \pls 1.6] & 1.00 & no \\
9B (1 seed) & AIME 2024 & 61.7$\to$59.6 & [\mns 7.9, \pls 3.8] & 1.00 & no \\
9B (1 seed) & AIME 2025 & 53.8$\to$49.2 & [\mns 8.8, \mns 0.4] & 0.34 & no \\
9B (1 seed) & GPQA-D & 42.9$\to$44.9 & [\mns 3.0, \pls 7.1] & 1.00 & no \\
9B (1 seed) & GSM8K & 93.7$\to$92.6 & [\mns 2.0, \mns 0.1] & 0.34 & yes \\
\rowcolor{keyrow} 9B (1 seed) & \textbf{macro} & \textbf{67.6$\to$66.3} & [\mns 3.1, \pls 0.5] & -- & no \\
\bottomrule
\end{tabular}
\end{table}

\para{Anchored recurrent depth results.} After post-training, ARD lowers accuracy. At depth 2 it scores 51.0\% against 60.3\% for its control ($p<0.001$, lower in both seeds), and accuracy falls with each extra pass, even on the training data. Before post-training, the loop does much less harm. At depth 2, ARD is close to its control (58.7\% vs 59.2\%, $p=0.50$), and at depth 1 it falls below it ($p<0.001$), so the model does use the second pass. The difference between the two timings is significant, with a 95\% interval of [\pls 6.3, \pls 11.2] points. Table~\ref{tab:ard} gives each setting at each test-time depth.

\begin{table}[t]
\centering
\caption{\textbf{ARD before or after post-training} (Qwen3.5-4B). Accuracy (\%) at test-time depth $K$. Each control uses the same adapter, data and steps, without the loop. AIME24/25 stands for AIME 2024/2025 and GPQA-D for GPQA-Diamond. $^\dagger$Joint re-evaluation. Shaded rows mark ARD after post-training at depth 2, and bold numbers are their macro accuracies, which fall well below those of the controls.}
\label{tab:ard}
\small
\setlength{\tabcolsep}{3pt}
\resizebox{\linewidth}{!}{%
\begin{tabular}{llcccccccc}
\toprule
Timing & Setting & Seed & $K$ & MATH-500 & AIME24 & AIME25 & GPQA-D & GSM8K & Macro \\
\midrule
before & A1: ARD continued pretraining $\to$ SFT$^\dagger$ & 0 & 1 & 78.4 & 47.9 & 35.0 & 36.4 & 89.6 & 57.5 \\
before & A1: ARD continued pretraining $\to$ SFT$^\dagger$ & 0 & 2 & 80.6 & 50.8 & 39.2 & 37.4 & 89.5 & 59.5 \\
before & A1: ARD continued pretraining $\to$ SFT & 1 & 1 & 77.6 & 44.6 & 37.9 & 27.8 & 87.9 & 55.1 \\
before & A1: ARD continued pretraining $\to$ SFT & 1 & 2 & 81.4 & 45.8 & 40.8 & 31.3 & 89.8 & 57.8 \\
before & A1c: control$^\dagger$ & 0 & 1 & 80.6 & 49.2 & 40.4 & 35.9 & 89.7 & 59.1 \\
before & A1c: control & 1 & 1 & 79.4 & 49.6 & 42.1 & 35.9 & 89.4 & 59.3 \\
before & A1c, FLOP-matched & 0 & 1 & 81.2 & 53.3 & 37.9 & 36.4 & 89.2 & 59.6 \\
after & A2: SFT $\to$ ARD$^\dagger$ & 0 & 1 & 79.6 & 46.2 & 39.2 & 32.3 & 89.6 & 57.4 \\
\rowcolor{keyrow} after & A2: SFT $\to$ ARD$^\dagger$ & 0 & 2 & 75.8 & 39.6 & 33.8 & 27.8 & 88.5 & \textbf{53.1} \\
after & A2: SFT $\to$ ARD & 1 & 1 & 79.8 & 46.2 & 36.7 & 39.4 & 88.9 & 58.2 \\
\rowcolor{keyrow} after & A2: SFT $\to$ ARD & 1 & 2 & 72.4 & 28.7 & 28.7 & 25.3 & 89.2 & \textbf{48.9} \\
after & A2c: control$^\dagger$ & 0 & 1 & 81.8 & 47.9 & 40.0 & 35.9 & 90.5 & 59.2 \\
after & A2c: control & 1 & 1 & 82.0 & 53.3 & 39.6 & 40.9 & 90.9 & 61.3 \\
after & A2c, FLOP-matched & 0 & 1 & 81.2 & 50.4 & 40.0 & 36.9 & 89.9 & 59.7 \\
\bottomrule
\end{tabular}}
\end{table}

% -----------------------------------------------------------------------------------------------------
\subsection{Equal FLOPs and decoding fixes}
\label{app:eqdetail}

Table~\ref{tab:eqc} in Section~\ref{sec:post} lists every equal-FLOP comparison.

Each looped evaluation has two possible comparators. The \emph{same-budget} comparator is the twin at the same token budget, which costs less than the loop. The \emph{best cheaper} comparator is the most accurate unlooped evaluation, over all budgets, whose FLOPs do not exceed the loop's. Choosing it on the test items favours the unlooped model.

\para{Short fine-tunes.} The looped 0.8B model leads the best cheaper evaluation at every budget from 0.5 to 2$\times$ and in measured wall-clock time (Figure~\ref{fig:frontier}). At 4B and 9B the loop is not significantly ahead at any budget. After Holm correction over the four budgets, the only remaining difference favours the unlooped model (9B at 0.5$\times$).

\para{Long fine-tunes.} The long 4B loop does not shorten its answers, and 10 of its 12 evaluations cost more than every unlooped one. Against the best cheaper evaluation, the loop is behind at 0.5$\times$ and 1$\times$ and level at 1.5$\times$. At 9B, a twin allowed 1.25$\times$ the generated tokens, at 1.16$\times$ the loop's time, beats the loop by four points (70.5\% vs 66.3\%, $p<0.001$).

\para{Decoding fixes.} We give the unlooped 0.8B model two decoding fixes that add no compute per token (Table~\ref{tab:fix}). One is budget forcing, and the other is a \emph{repetition stop} that ends the reasoning when the output repeats a period of 4--256 tokens three times. The repetition stop gains less, so we use budget forcing as our baseline. Budget forcing on the twin alone beats the plain looped model at no greater cost, at 0.8B as at 4B and 9B. With budget forcing on both 0.8B models, the loop leads again (four seeds), but only on GSM8K ($p<0.001$, against $p=0.77$ on MATH-500). This gain depends on the tokens reserved for the answer. With 1{,}024 instead of 512 reserved tokens (seed 0), it is not significant at the default budget ($p=0.24$) but remains at larger budgets ($p\le0.005$).

\para{More fine-tuning.} Training is a one-off cost, so we also fine-tune the 0.8B twin on more tokens (seeds 0 and 1). With plain decoding, twins fine-tuned on 1.5$\times$ and 2$\times$ the tokens still trail the looped model ($p<0.001$), and only 3$\times$ overtakes it (50.4\% vs 48.6\%, $p=0.04$). With budget forcing on both, the 3$\times$ twin beats the 1$\times$ looped model (60.1\% vs 56.5\%, $p<0.001$). Given the same 3$\times$ fine-tuning, however, the looped model leads again (64.1\% vs 60.1\%, $p<0.001$), in both seeds and at every budget. Training only the twin longer is therefore not a fair test of the loop.

% -----------------------------------------------------------------------------------------------------
\subsection{Serving}
\label{app:serving}

\begin{table}[htbp]
\centering
\caption{\textbf{Serving cost} (one H100, unlooped $\to$ looped). Latency is the decode time per token at batch 1. We measure throughput at the largest batch that fits in 80\,GB, and memory as the peak at batch 1. Bold ratios are the key results.}
\label{tab:serve}
\small
\begin{tabular}{lcccccc}
\toprule
Model & Latency (ms) & ($\times$) & Throughput (tokens/s) & ($\times$) & Batch & Memory (GB) \\
\midrule
0.8B & 20.6 $\to$ 25.3 & \textbf{1.23} & 6{,}842 $\to$ 5{,}286 & \textbf{0.77} & 693 $\to$ 710 & 1.5 $\to$ 1.5 \\
4B & 29.2 $\to$ 36.8 & \textbf{1.26} & 2{,}552 $\to$ 2{,}026 & \textbf{0.79} & 324 $\to$ 298 & 8.1 $\to$ 8.1 \\
9B & 29.4 $\to$ 36.0 & \textbf{1.22} & 2{,}400 $\to$ 1{,}923 & \textbf{0.80} & 245 $\to$ 247 & 16.9 $\to$ 17.0 \\
\bottomrule
\end{tabular}
\end{table}

Table~\ref{tab:serve} gives the serving cost of each loop. The loop slows every decoding step and enlarges every KV cache, so it lowers throughput as well as raising latency. Per token, the looped model's KV cache is 1.33$\times$ (0.8B) and 1.25$\times$ (4B, 9B) that of its twin.

% =====================================================================================================
\section{More training through the loop}
\label{app:sec4}

% -----------------------------------------------------------------------------------------------------
\subsection{Continued pretraining}

We continue pretraining Qwen3.5-0.8B (two seeds) and 4B (one seed) with and without the loop, and measure held-out loss on pretraining text (Figure~\ref{fig:sweep}). The loop passes three thresholds at very different speeds. First, it learns to \emph{use} its second pass quickly, and within 25--50M tokens the second pass lowers its own loss (green curve in Figure~\ref{fig:sweep}). Secondly, it \emph{beats its twin} later, from about 0.5B tokens onwards, in both 0.8B seeds and the 4B seed (blue curve). Thirdly, it never \emph{beats equal training compute} where we can measure it (red curve). At 5B tokens, the looped 0.8B model is still behind a twin trained on 1.25$\times$ the tokens.

% -----------------------------------------------------------------------------------------------------
\subsection{Token exchange rate}

The token exchange rate $r(D)$ is the number of tokens the twin needs to match the looped model after $D$ tokens, divided by $D$. The loop pays for its training compute only if $r$ exceeds its training cost, which is 1.25 counting layers, 1.16 counting FLOPs including the output head, and 1.19 in measured training time.

\begin{figure}[t]
\centering
\includegraphics[width=0.62\linewidth]{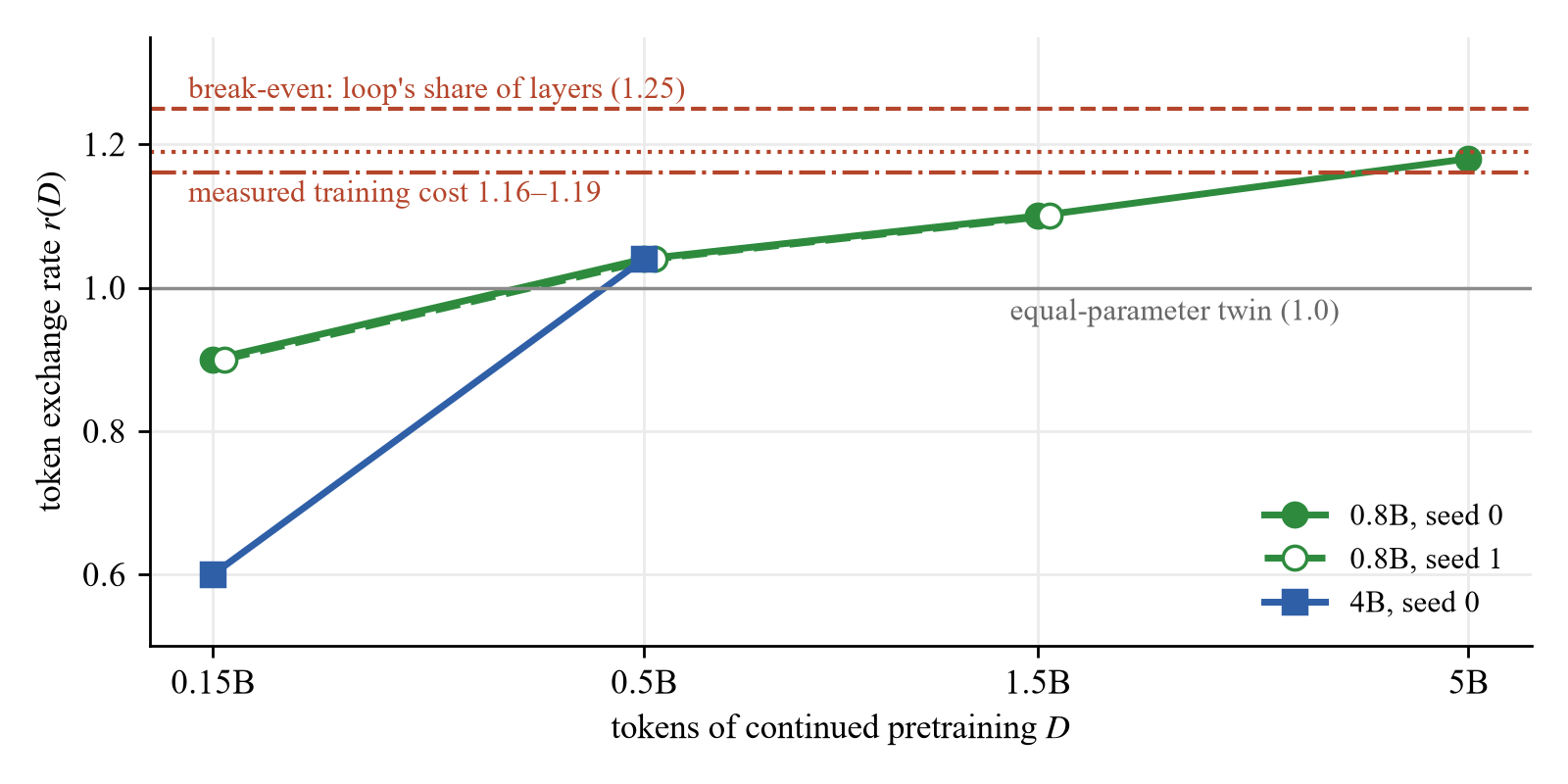}
\caption{\textbf{Token exchange rate} $r(D)$, i.e.\ how many tokens the twin needs per looped-model token to reach the same held-out loss. Above 1 (grey line), the loop beats its twin. Above the red lines, it also pays for its training compute, which is 1.25 counting layers and 1.16--1.19 as measured. At 5B tokens, the 0.8B loop (1.18) passes the lowest cost estimate but not the others.}
\label{fig:exr}
\end{figure}

For 0.8B, $r$ rises from 0.90 at 0.15B tokens to 1.10 at 1.5B and 1.18 at 5B (Figure~\ref{fig:exr}). The loop is therefore within a few percent of paying for its training compute, and further from paying for its inference compute, which it adds on every served token.

% -----------------------------------------------------------------------------------------------------
\subsection{From loss to accuracy}

We apply the short reasoning fine-tune to the 0.8B checkpoints after 5.2B tokens (two seeds). The looped model reaches 37.6\% macro accuracy, against 31.6\% for its twin ($p<0.001$), and it also leads at equal FLOPs (Table~\ref{tab:eqc}). Two caveats apply. Continued pretraining on our web-and-math mix lowers the fine-tuned accuracy of every model, and the loop loses less, so part of its lead is protection from this degradation. Moreover, every continued-pretraining model stays below the plain fine-tune of the base model (46.4\% for the best against 54.7\%), so continued pretraining on our mix does not help deployment.

% -----------------------------------------------------------------------------------------------------
\subsection{Loops learned from scratch}
\label{app:pretrainfull}
\label{sec:pretrain}

To separate looping itself from adding the loop late, we train a 113M-parameter Llama-style model from scratch on 1.5B tokens of FineWeb-Edu, with the same data order, optimiser and schedule for every model. The looped model runs layers 4--7 twice with the damped update (1.22$\times$ the FLOPs per token). Table~\ref{tab:pretrain} in Section~\ref{sec:budget} compares it with a dense model of the same training FLOPs, a 16-layer dense model, loops added after pretraining, random depth (the core applied a random number of times per batch) and an input-injected core~\cite{geiping2025scaling}. We report held-out loss and zero-shot HellaSwag~\cite{zellers2019hellaswag}, ARC-Easy and ARC-Challenge~\cite{clark2018arc} and LAMBADA~\cite{paperno2016lambada}.

Section~\ref{sec:budget} discusses the results. Downstream accuracy follows the same order as the held-out loss.

% =====================================================================================================
\section{Training-free loops on released models}
\label{app:released}

We name two Qwen3-8B structures. \emph{G-best}, the best loop of a grid search, loops layers 13--19 twice ($\beta=0.25$) with a threefold MLP loop at layer 16 (cost 0.25). \emph{S-noattn}, the loop found by our full search without its attention-only loops, loops layers 14--20 twice ($\beta=0.25$) with an MLP loop at layer 16 (cost 0.25), and we use it for conversion.

% -----------------------------------------------------------------------------------------------------
\subsection{Where calibrated search places the loop}

\para{An exhaustive grid suffices.} On Qwen3-8B we score all 2{,}016 single-window configurations by $\pplT$, covering every start, widths 2--8, $K\in\{2,3,4\}$ and every anchor. Adding one MLP loop inside the best windows gives G-best. The window found by our full search ranks 16th of the 2{,}016 grid configurations, so a simple grid is enough to find the useful location.

\begin{figure}[t]
\centering
\begin{minipage}{0.49\linewidth}\centering\includegraphics[width=\linewidth]{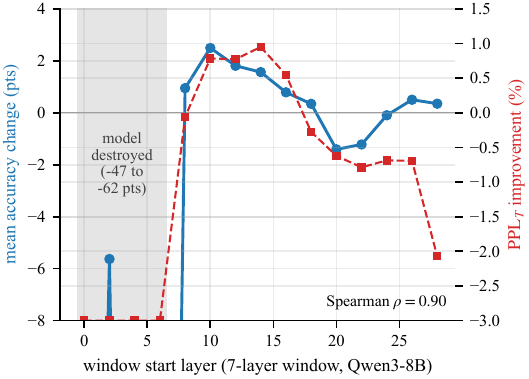}\end{minipage}\hfill
\begin{minipage}{0.49\linewidth}\centering\includegraphics[width=\linewidth]{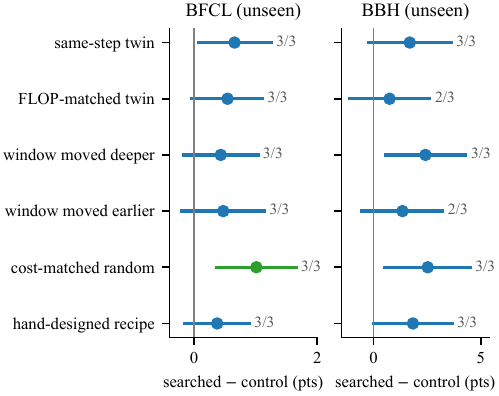}\end{minipage}
\caption{\textbf{Where to loop, and what conversion keeps (Qwen3-8B).} \emph{Left}, a seven-layer window moves through the network. Blue shows the change in mean accuracy on GSM8K, MATH-500 and MMLU-Pro relative to the unmodified model, red the $\pplT$ improvement, and shading marks windows that destroy the model. Spearman $\rho$ is the rank correlation between the two curves across the 17 placements. \emph{Right}, the searched placement minus each control after the same conversion, with 95\% intervals over three seeds, on two benchmarks never used for selection, the Berkeley Function Calling Leaderboard (BFCL) and BIG-Bench Hard (BBH). The controls are the same-step twin, a twin trained for the loop's training FLOPs, the same window moved deeper or earlier, a random structure of the same cost, and the hand-designed loop recipe of Chen et~al.~\cite{chen2026training} at matched cost. Labels such as 3/3 count the seeds in which the searched placement is ahead, and green marks significance after Holm correction.}
\label{fig:main}
\end{figure}

\para{Location profile.} Figure~\ref{fig:main} (left) moves a seven-layer window ($K=2$, $\beta=0.25$) through the network in steps of two layers, giving 17 placements of identical cost. Accuracy on GSM8K, MATH-500 and MMLU-Pro traces an inverted U over depth. Windows starting in the first seven layers destroy the model, windows starting at layers 8--16 help, and deeper ones are neutral or harmful. $\pplT$ ranks the 17 placements in the same order as their benchmark mean (Spearman $\rho=0.90$), and the 13 non-destructive ones too ($\rho=0.79$, against $\rho=0.66$ for plain perplexity).

\para{The band is stable.} Across 26 searches with different seeds, objectives and models, the best structures centre their window at 0.47$\pm$0.03 of depth. This agrees with the depth rule that Chen et~al.~\cite{chen2026training} find with benchmark labels (0.45--0.60 of depth), and we recover it without labels.

% -----------------------------------------------------------------------------------------------------
\subsection{Downstream results}
\label{app:full}

\para{Qwen3-8B.} G-best raises GSM8K and MMLU-Pro in each of three replicate evaluations of the unmodified model, while its change on MATH-500 lies within run-to-run variation. Undamped repetition of the same seven layers cuts accuracy sharply (MATH-500 54.0\% vs 76.3\%), so damping, not repetition, makes reuse of these layers safe.

\para{Across models.} On Qwen3-8B, the best calibrated configuration significantly improves GSM8K and MMLU-Pro at an estimated 1.74$\times$ decoding slowdown. At 1.7B and 14B, changes are small and mostly not significant. On Qwen3-32B, the winner is higher on all six tasks we evaluate, but only by 0.4--1.7 points, at 2.9$\times$ the evaluation time. Training-free looping therefore helps some models on some tasks, but not reliably.

\para{Measured cost and cheaper alternatives.} With batch size 8 on one GPU, the best calibrated 8B configuration decodes 1.80$\times$ slower than the unmodified model. Unmodified Qwen3-14B decodes faster than the looped 8B model and beats it on MATH-500 and MMLU-Pro, though not on GSM8K. On AIME 2024 and thinking-mode GSM8K, looping and self-consistency (a majority vote over 2 or 4 samples) do not differ significantly (Table~\ref{tab:cc}). On MATH-500, looping adds less than one point at 1.7$\times$ the time, while self-consistency and Qwen3-14B each add four to five points. Looping a released model is therefore only useful when weight memory, and not decoding time, is the limiting constraint.

\begin{table}[htbp]
\centering
\caption{\textbf{Looping versus repeated sampling.} Accuracy (\%) and time relative to greedy decoding of the unmodified model. Looped PPL$_T$ marks the loop chosen by calibrated perplexity. Shaded rows mark the key results.}
\label{tab:cc}
\small
\begin{tabular}{lcccc}
\toprule
& \multicolumn{2}{c}{AIME 2024} & \multicolumn{2}{c}{GSM8K (thinking)} \\
\cmidrule(lr){2-3}\cmidrule(lr){4-5}
Method & acc. & time & acc. & time \\
\midrule
\multicolumn{5}{l}{\textit{Qwen3-1.7B}} \\
\quad greedy & 16.7 & 1.00$\times$ & 81.0 & 1.00$\times$ \\
\quad self-consistency, 2 samples & 13.3 & 2.01$\times$ & 83.0 & 1.34$\times$ \\
\quad self-consistency, 4 samples & 26.7 & 3.23$\times$ & 85.0 & 1.90$\times$ \\
\quad looped, PPL$_T$ & 13.3 & 2.43$\times$ & 84.0 & 1.75$\times$ \\
\quad looped, PPL$_T$, $\le$25\% cost & 16.7 & 0.73$\times$ & 84.0 & 1.13$\times$ \\
\midrule
\multicolumn{5}{l}{\textit{Qwen3-8B}} \\
\quad greedy & 26.7 & 1.00$\times$ & 86.0 & 1.00$\times$ \\
\quad self-consistency, 2 samples & 26.7 & 1.56$\times$ & 85.0 & 1.73$\times$ \\
\rowcolor{keyrow} \quad self-consistency, 4 samples & 33.3 & 2.47$\times$ & 86.0 & 3.01$\times$ \\
\rowcolor{keyrow} \quad looped, PPL$_T$ & 30.0 & 1.71$\times$ & 87.0 & 1.88$\times$ \\
\quad looped, PPL$_T$, $\le$25\% cost & 26.7 & 1.02$\times$ & 89.0 & 1.25$\times$ \\
\bottomrule
\end{tabular}
\end{table}

% -----------------------------------------------------------------------------------------------------
\subsection{Conversion}
\label{appx:conv}

Does training through the loop improve the training-free result? We test this by conversion and report the \emph{structure effect}, i.e.\ the converted looped model minus its identically trained twin.

\para{Training shrinks the loop's contribution.} Without training, S-noattn beats the unmodified model on all three benchmarks (GSM8K 92.0\% vs 88.7\%). After conversion, it is within one point of its twin on every benchmark (mean of three seeds), because conversion raises both models about equally. What the loop added is therefore largely what light training also gives.

\para{Placement stops mattering after conversion.} After the same conversion, the searched structure is no longer clearly better than relocated windows, a random structure or a hand-designed recipe on GSM8K, MATH-500 and MMLU-Pro. On two benchmarks never used for selection, tool calling on the Berkeley Function Calling Leaderboard (BFCL v3)~\cite{patil2025bfcl} and BIG-Bench Hard (BBH)~\cite{suzgun2023bbh}, it leads every control, but only the gap to the random structure is significant (Figure~\ref{fig:main}, right).

% =====================================================================================================
\section{Calibrated perplexity as a search objective}
\label{sec:search}

\para{Why a search objective matters.} To place a loop without using benchmark answers, we need a cheap score that predicts which placements help. Neural architecture search relies on such cheap proxies~\cite{real2019regularized,liu2019darts,abdelfattah2021zero,ning2021evaluating}. Perplexity on held-out text is the obvious choice, but it fails here because post-trained language models are over-confident, as modern neural networks often are~\cite{guo2017calibration}. This appendix shows the failure and the fix.

\begin{figure}[htbp]
\centering
\includegraphics[width=\linewidth]{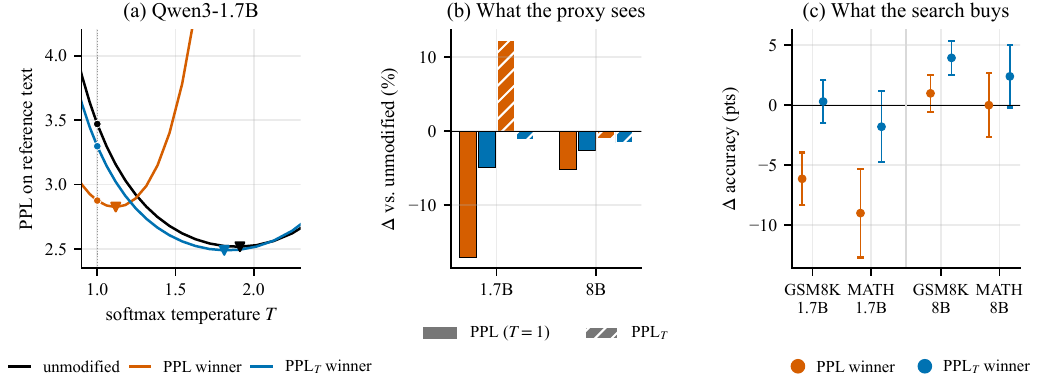}
\caption{\textbf{Why we place loops by calibrated perplexity.} Orange shows the loop picked by plain perplexity (PPL), blue the loop picked by calibrated perplexity (PPL$_T$), and black the unmodified model. (a)~Perplexity against softmax temperature on Qwen3-1.7B. Circles mark $T=1$ and triangles each model's best temperature. The orange loop looks good at $T=1$ only because it flattens an over-confident model, and the unmodified model reaches a lower perplexity by raising its temperature alone. (b)~Change in perplexity relative to the unmodified model, at $T=1$ (solid) and at each model's best temperature (hatched). The orange loop's apparent gain becomes a loss once temperature is taken into account. (c)~Change in accuracy with 95\% intervals. The PPL winner hurts accuracy at 1.7B, while the PPL$_T$ winner does not.}
\label{fig:hero}
\end{figure}

\para{Plain perplexity rediscovers temperature scaling.} A post-trained model puts too much probability on its top token, so its perplexity drops if we simply soften its outputs by raising the softmax temperature. A loop can lower perplexity in the same way, without making the model any better. On Qwen3-1.7B, a search on plain perplexity finds a loop that lowers perplexity by 17.1\% (Figure~\ref{fig:hero}b, solid orange bar). Yet raising the temperature of the unmodified model, with no loop at all, lowers its perplexity by 27.4\% (Figure~\ref{fig:hero}a, black curve). When we compare both models at their own best temperature, the searched loop is 12.2\% worse (hatched orange bar), and it is much less accurate, for example 59.2\% against 68.2\% on MATH-500 (Figure~\ref{fig:hero}c). We therefore search on calibrated perplexity,
\begin{equation}
\pplT(e)=\exp\min_{T>0}L_e(T),
\end{equation}
where $L_e(T)$ is the mean token loss of candidate $e$ when its logits are divided by $T$. This is the perplexity at the candidate's own best temperature. Rescaling the logits cannot improve $\pplT$, so a loop scores well only if it predicts the text better. The loops chosen this way (blue in Figure~\ref{fig:hero}) do not hurt accuracy, and at 8B they improve GSM8K by about four points.

\para{A better score does not guarantee better accuracy.} Calibrated perplexity avoids this trap, but it is not a perfect guide. The loops it selects improve $\pplT$ at every model size from 1.7B to 32B, yet their benchmark accuracy improves clearly only at 8B, slightly at 32B, and not at 1.7B or 14B (Appendix~\ref{app:full}). The score therefore rules out harmful placements, but it does not promise a gain. This matches the main finding that training-free loops help only some models on some tasks.

\para{Search algorithms.} Local search and regularised evolution~\cite{real2019regularized} are the most reliable of the nine search algorithms we compare, and they find the same mid-depth band. The choice of algorithm matters much less than the choice of objective.

\end{document}